\documentclass[conference,transmag]{IEEEtran}
\usepackage{amsmath,amssymb,amsfonts}
\usepackage{algorithm}
\usepackage{algpseudocode}
\usepackage{lipsum}
\usepackage{float}
\usepackage{cite}
\usepackage{caption}
\usepackage{subcaption}
\usepackage{comment}
\usepackage{adjustbox}

\usepackage{xcolor}
\usepackage{graphicx}
\usepackage{acronym}

\usepackage{fixltx2e}

\graphicspath{{images/}} 
\usepackage{todonotes}

\begin{document}

\acrodef{ADC}[ADC]{Analog to Digital Converter}
\acrodef{ADEXP}[AdExp-I\&F]{Adaptive-Exponential Integrate and Fire}
\acrodef{ADM}[ADM]{Asynchronous Delta Modulator}
\acrodef{AER}[AER]{Address-Event Representation}
\acrodef{AEX}[AEX]{AER EXtension board}
\acrodef{AE}[AE]{Address-Event}
\acrodef{AFM}[AFM]{Atomic Force Microscope}
\acrodef{AGC}[AGC]{Automatic Gain Control}
\acrodef{AI}[AI]{Artificial Intelligence}
\acrodef{AMDA}[AMDA]{AER Motherboard with D/A converters}
\acrodef{ANN}[ANN]{Artificial Neural Network}
\acrodef{API}[API]{Application Programming Interface}
\acrodef{APMOM}[APMOM]{Alternate Polarity Metal On Metal}
\acrodef{ARM}[ARM]{Advanced RISC Machine}
\acrodef{ASIC}[ASIC]{Application Specific Integrated Circuit}
\acrodef{AdExp}[AdExp-IF]{Adaptive Exponential Integrate-and-Fire}
\acrodef{BCM}[BMC]{Bienenstock-Cooper-Munro}
\acrodef{BD}[BD]{Bundled Data}
\acrodef{BEOL}[BEOL]{Back-end of Line}
\acrodef{BG}[BG]{Bias Generator}
\acrodef{BMI}[BMI]{Brain-Machince Interface}
\acrodef{BTB}[BTB]{band-to-band tunnelling}
\acrodef{CAD}[CAD]{Computer Aided Design}
\acrodef{CAM}[CAM]{Content Addressable Memory}
\acrodef{CAVIAR}[CAVIAR]{Convolution AER Vision Architecture for Real-Time}
\acrodef{CA}[CA]{Cortical Automaton}
\acrodef{CCN}[CCN]{Cooperative and Competitive Network}
\acrodef{CDR}[CDR]{Clock-Data Recovery}
\acrodef{CFC}[CFC]{Current to Frequency Converter}
\acrodef{CHP}[CHP]{Communicating Hardware Processes}
\acrodef{CMIM}[CMIM]{Metal-insulator-metal Capacitor}
\acrodef{CML}[CML]{Current Mode Logic}
\acrodef{CMOL}[CMOL]{Hybrid CMOS nanoelectronic circuits}
\acrodef{CMOS}[CMOS]{Complementary Metal-Oxide-Semiconductor}
\acrodef{CNN}[CCN]{Convolutional Neural Network}
\acrodef{COTS}[COTS]{Commercial Off-The-Shelf}
\acrodef{CPG}[CPG]{Central Pattern Generator}
\acrodef{CPLD}[CPLD]{Complex Programmable Logic Device}
\acrodef{CPU}[CPU]{Central Processing Unit}
\acrodef{CSM}[CSM]{Cortical State Machine}
\acrodef{CSP}[CSP]{Constraint Satisfaction Problem}
\acrodef{CTXCTL}[CTXCTL]{CortexControl}
\acrodef{CV}[CV]{Coefficient of Variation}
\acrodef{DAC}[DAC]{Digital to Analog Converter}
\acrodef{DAS}[DAS]{Dynamic Auditory Sensor}
\acrodef{DAVIS}[DAVIS]{Dynamic and Active Pixel Vision Sensor}
\acrodef{DBN}[DBN]{Deep Belief Network}
\acrodef{DFA}[DFA]{Deterministic Finite Automaton}
\acrodef{DIBL}[DIBL]{drain-induced-barrier-lowering}
\acrodef{DI}[DI]{delay insensitive}
\acrodef{DMA}[DMA]{Direct Memory Access}
\acrodef{DNF}[DNF]{Dynamic Neural Field}
\acrodef{DNN}[DNN]{Deep Neural Network}
\acrodef{DOF}[DOF]{Degrees of Freedom}
\acrodef{DPE}[DPE]{Dynamic Parameter Estimation}
\acrodef{DPI}[DPI]{Differential Pair Integrator}
\acrodef{DRAM}[DRAM]{Dynamic Random Access Memory}
\acrodef{DRRZ}[DR-RZ]{Dual-Rail Return-to-Zero}
\acrodef{DR}[DR]{Dual Rail}
\acrodef{DSP}[DSP]{Digital Signal Processor}
\acrodef{DVS}[DVS]{Dynamic Vision Sensor}
\acrodef{DYNAP}[DYNAP]{Dynamic Neuromorphic Asynchronous Processor}
\acrodef{EBL}[EBL]{Electron Beam Lithography}
\acrodef{EDVAC}[EDVAC]{Electronic Discrete Variable Automatic Computer}
\acrodef{EEG}[EEG]{electroencephalography}
\acrodef{EIN}[EIN]{Excitatory-Inhibitory Network}
\acrodef{EM}[EM]{Expectation Maximization}
\acrodef{EPSC}[EPSC]{Excitatory Post-Synaptic Current}
\acrodef{EPSP}[EPSP]{Excitatory Post-Synaptic Potential}
\acrodef{EZ}[EZ]{Epileptogenic Zone}
\acrodef{FDSOI}[FDSOI]{Fully-Depleted Silicon on Insulator}
\acrodef{FET}[FET]{Field-Effect Transistor}
\acrodef{FFT}[FFT]{Fast Fourier Transform}
\acrodef{FI}[F-I]{Frequency-Current}
\acrodef{FPGA}[FPGA]{Field Programmable Gate Array}
\acrodef{FR}[FR]{Fast Ripple}
\acrodef{FSA}[FSA]{Finite State Automaton}
\acrodef{FSM}[FSM]{Finite State Machine}
\acrodef{GIDL}[GIDL]{gate-induced-drain-leakage}
\acrodef{GOPS}[GOPS]{Giga-Operations per Second}
\acrodef{GPU}[GPU]{Graphical Processing Unit}
\acrodef{GUI}[GUI]{Graphical User Interface}
\acrodef{HAL}[HAL]{Hardware Abstraction Layer}
\acrodef{HFO}[HFO]{High Frequency Oscillation}
\acrodef{HH}[H\&H]{Hodgkin \& Huxley}
\acrodef{HMM}[HMM]{Hidden Markov Model}
\acrodef{HRS}[HRS]{High-Resistive State}
\acrodef{HR}[HR]{Human Readable}
\acrodef{HSE}[HSE]{Handshaking Expansion}
\acrodef{HW}[HW]{Hardware}
\acrodef{ICT}[ICT]{Information and Communication Technology}
\acrodef{IC}[IC]{Integrated Circuit}
\acrodef{IEEG}[iEEG]{intracranial electroencephalography}
\acrodef{IF2DWTA}[IF2DWTA]{Integrate \& Fire 2--Dimensional WTA}
\acrodef{IFSLWTA}[IFSLWTA]{Integrate \& Fire Stop Learning WTA}
\acrodef{IF}[I\&F]{Integrate-and-Fire}
\acrodef{IMU}[IMU]{Inertial Measurement Unit}
\acrodef{INCF}[INCF]{International Neuroinformatics Coordinating Facility}
\acrodef{INI}[INI]{Institute of Neuroinformatics}
\acrodef{IO}[I/O]{Input/Output}
\acrodef{IPSC}[IPSC]{Inhibitory Post-Synaptic Current}
\acrodef{IPSP}[IPSP]{Inhibitory Post-Synaptic Potential}
\acrodef{IP}[IP]{Intellectual Property}
\acrodef{ISI}[ISI]{Inter-Spike Interval}
\acrodef{IoT}[IoT]{Internet of Things}
\acrodef{JFLAP}[JFLAP]{Java - Formal Languages and Automata Package}
\acrodef{LEDR}[LEDR]{Level-Encoded Dual-Rail}
\acrodef{LFP}[LFP]{Local Field Potential}
\acrodef{LLC}[LLC]{Low Leakage Cell}
\acrodef{LNA}[LNA]{Low-Noise Amplifier}
\acrodef{LPF}[LPF]{Low Pass Filter}
\acrodef{LRS}[LRS]{Low-Resistive State}
\acrodef{LSM}[LSM]{Liquid State Machine}
\acrodef{LTD}[LTD]{Long Term Depression}
\acrodef{LTI}[LTI]{Linear Time-Invariant}
\acrodef{LTP}[LTP]{Long Term Potentiation}
\acrodef{LTU}[LTU]{Linear Threshold Unit}
\acrodef{LUT}[LUT]{Look-Up Table}
\acrodef{LVDS}[LVDS]{Low Voltage Differential Signaling}
\acrodef{MCMC}[MCMC]{Markov-Chain Monte Carlo}
\acrodef{MEMS}[MEMS]{Micro Electro Mechanical System}
\acrodef{MFR}[MFR]{Mean Firing Rate}
\acrodef{MIM}[MIM]{Metal Insulator Metal}
\acrodef{MLP}[MLP]{Multilayer Perceptron}
\acrodef{MOSCAP}[MOSCAP]{Metal Oxide Semiconductor Capacitor}
\acrodef{MOSFET}[MOSFET]{Metal Oxide Semiconductor Field-Effect Transistor}
\acrodef{MOS}[MOS]{Metal Oxide Semiconductor}
\acrodef{MRI}[MRI]{Magnetic Resonance Imaging}
\acrodef{NDFSM}[NDFSM]{Non-deterministic Finite State Machine} 
\acrodef{ND}[ND]{Noise-Driven}
\acrodef{NEF}[NEF]{Neural Engineering Framework}
\acrodef{NHML}[NHML]{Neuromorphic Hardware Mark-up Language}
\acrodef{NIL}[NIL]{Nano-Imprint Lithography}
\acrodef{NMDA}[NMDA]{N-Methyl-D-Aspartate}
\acrodef{NME}[NE]{Neuromorphic Engineering}
\acrodef{NN}[NN]{Neural Network}
\acrodef{NOC}[NoC]{Network-on-Chip}
\acrodef{NRZ}[NRZ]{Non-Return-to-Zero}
\acrodef{NSM}[NSM]{Neural State Machine}
\acrodef{OR}[OR]{Operating Room}
\acrodef{OTA}[OTA]{Operational Transconductance Amplifier}
\acrodef{PCB}[PCB]{Printed Circuit Board}
\acrodef{PCHB}[PCHB]{Pre-Charge Half-Buffer}
\acrodef{PCM}[PCM]{Phase Change Memory}
\acrodef{PE}[PE]{Phase Encoding}
\acrodef{PFA}[PFA]{Probabilistic Finite Automaton}
\acrodef{PFC}[PFC]{prefrontal cortex}
\acrodef{PFM}[PFM]{Pulse Frequency Modulation}
\acrodef{PR}[PR]{Production Rule}
\acrodef{PSC}[PSC]{Post-Synaptic Current}
\acrodef{PSP}[PSP]{Post-Synaptic Potential}
\acrodef{PSTH}[PSTH]{Peri-Stimulus Time Histogram}
\acrodef{QDI}[QDI]{Quasi Delay Insensitive}
\acrodef{RAM}[RAM]{Random Access Memory}
\acrodef{RA}[RA]{Resected Area}
\acrodef{RDF}[RDF]{random dopant fluctuation}
\acrodef{RELU}[ReLu]{Rectified Linear Unit}
\acrodef{RLS}[RLS]{Recursive Least-Squares}
\acrodef{RMSE}[RMSE]{Root Mean Squared-Error}
\acrodef{RMS}[RMS]{Root Mean Squared}
\acrodef{RNN}[RNN]{Recurrent Neural Networks}
\acrodef{ROLLS}[ROLLS]{Reconfigurable On-Line Learning Spiking}
\acrodef{RRAM}[R-RAM]{Resistive Random Access Memory}
\acrodef{R}[R]{Ripples}
\acrodef{SAC}[SAC]{Selective Attention Chip}
\acrodef{SAT}[SAT]{Boolean Satisfiability Problem}
\acrodef{SCX}[SCX]{Silicon CorteX}
\acrodef{SD}[SD]{Signal-Driven}
\acrodef{SEM}[SEM]{Spike-based Expectation Maximization}
\acrodef{SLAM}[SLAM]{Simultaneous Localization and Mapping}
\acrodef{SNN}[SNN]{Spiking Neural Network}
\acrodef{SNR}[SNR]{Signal to Noise Ratio}
\acrodef{SOC}[SOC]{System-On-Chip}
\acrodef{SOI}[SOI]{Silicon on Insulator}
\acrodef{SOZ}[SOZ]{Seizure Onset Zone}
\acrodef{SP}[SP]{Separation Property}
\acrodef{SRAM}[SRAM]{Static Random Access Memory}
\acrodef{STDP}[STDP]{Spike-Timing Dependent Plasticity}
\acrodef{STD}[STD]{Short-Term Depression}
\acrodef{STP}[STP]{Short-Term Plasticity}
\acrodef{STT-MRAM}[STT-MRAM]{Spin-Transfer Torque Magnetic Random Access Memory}
\acrodef{STT}[STT]{Spin-Transfer Torque}
\acrodef{SW}[SW]{Software}
\acrodef{TCAM}[TCAM]{Ternary Content-Addressable Memory}
\acrodef{TFT}[TFT]{Thin Film Transistor}
\acrodef{TLE}[TLE]{Temporal Lobe Epilepsy}
\acrodef{USB}[USB]{Universal Serial Bus}
\acrodef{VHDL}[VHDL]{VHSIC Hardware Description Language}
\acrodef{VLSI}[VLSI]{Very Large Scale Integration}
\acrodef{VOR}[VOR]{Vestibulo-Ocular Reflex}
\acrodef{WCST}[WCST]{Wisconsin Card Sorting Test}
\acrodef{WTA}[WTA]{Winner-Take-All}
\acrodef{XML}[XML]{eXtensible Mark-up Language}
\acrodef{divmod3}[DIVMOD3]{divisibility of a number by three}
\acrodef{hWTA}[hWTA]{hard Winner-Take-All}
\acrodef{sWTA}[sWTA]{soft Winner-Take-All}

\title{A hardware-software co-design approach to minimize the use of memory resources in multi-core neuromorphic processors}

\author{\IEEEauthorblockN{
Vanessa R. C. Leite,
Zhe Su,
Adrian M. Whatley,
Giacomo Indiveri}
\IEEEauthorblockA{
Institute of Neuroinformatics,
University of Zurich and ETH Zurich,\\
Winterthurerstr. 190, 8057 Zurich, Switzerland\\
Email: vanessa@ini.uzh.ch}
}

\maketitle

\begin{abstract}
  Both in electronics and biology, physical implementations of neural networks have severe energy and memory constraints.
  We propose a hardware-software co-design approach for minimizing the use of memory resources in multi-core neuromorphic processors, by taking  inspiration from biological neural networks.
  We use this approach to design new routing schemes optimized for small-world networks and to provide guidelines for designing novel application-specific multi-core neuromorphic chips.
  Starting from the hierarchical routing scheme proposed, we present a hardware-aware placement algorithm that optimizes the allocation of resources for arbitrary network models.
  We validate the algorithm with a canonical small-world network and present preliminary results for other networks derived from it.
\end{abstract}

\begin{IEEEkeywords}
compiler, neuromorphic processors, hardware-software co-design, hierarchical routing, small-world networks, multi-core, scaling
\end{IEEEkeywords}

\section{Introduction}
\label{sec:introduction}
The large energy costs of \ac{DNN} and \ac{AI} algorithms are pushing the development of domain-specific hardware accelerators~\cite{Ibtesam_etal21}.
Neuromorphic processors are a class of \ac{AI} hardware accelerators that implement computational models of \acp{SNN} adopting in-memory computing strategies and brain-inspired principles of computation~\cite{Roy_etal19,Chicca_Indiveri20,Sebastian_etal20}.
They represent a very promising approach, especially for edge-computing applications, as they have the potential to reduce power consumption to ultra-low (e.g., sub milliwatt) figures~\cite{Covi_etal21}.
However, the requirement of \ac{SNN} hardware accelerators to store the state of each neuron, combined with their in-memory computing circuit design techniques leads to very large area consumption figures, which limits the sizes and numbers of parameters of the networks that they can implement.
The current strategy used to support the integration of large \ac{SNN} models in these accelerators is to use multi-core architectures~\cite{Moradi_etal18,Akopyan_etal15,Davies_etal18,Furber_Bogdan20}.
In these architectures, each core either \emph{emulates} with analog circuits~\cite{Moradi_etal18} or \emph{simulates} with time-multiplexed digital circuits~\cite{Akopyan_etal15,Davies_etal18,Furber_Bogdan20}, neuro-synaptic arrays in which both the synaptic weight matrix and the network connectivity routing memory blocks occupy a significant proportion of the total layout area.
Although the advent of nano-scale memristive devices can mitigate this problem by enabling the construction of dense cross-bar array structures for storing the weight matrices~\cite{Sebastian_etal20}, the problem of allocating routing and connectivity resources to allow arbitrary networks (e.g., with all-to-all possible connections) at scale is of a fundamental nature that even memristors or 3D-VLSI technologies cannot solve~\cite{Laughlin_Sejnowski03}.

Finding trade-offs to optimize both weight-matrix and connectivity/routing memory structures in multi-core neuromorphic processors can therefore have a significant impact on their total chip die area and on the size of the networks they can implement.
Following the original neuromorphic engineering approach~\cite{Mead90}, in this paper, we look at animal brains for inspiration and propose brain-inspired strategies to perform this optimization.
Specifically, we show that, by adopting small-world type network size/connectivity, we implement trade-offs that minimize area consumption requirements while still enabling the design of \ac{SNN} architectures that can solve a wide range of relevant ``edge-computing'' problems, i.e., the types of sensory-motor processing problems that animals solve in the real world.

\begin{figure}
  \centering
  \includegraphics[width=0.3\textwidth]{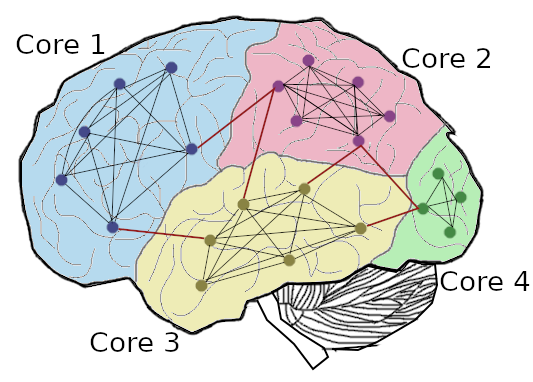}
  \caption{Small-world network connectivity in the brain.}
  \label{fig:small-world-network}
\end{figure}

\section{Neural network connectivity schemes}
\label{sec:nn-connectivity}

\subsection{In biological systems}
In animal brains, computation and other functions emerge from the interaction of neural areas.
The diverse patterns of connections found within and across different brain areas are highly correlated with brain functions such as memory, vision and motor control~\cite{Lynn_Bassett19}.
Recent studies relating structure to function show how the wiring of the brain is far from homogeneous~\cite{Lynn_Bassett19}: brain networks have short path length, high clustering, and a modular community structure~\cite{Bullmore_Sporns09}.
These biological neural networks are often highly recurrent and have dense connections among nearby neurons and sparse connections to specific/far-away neurons, showing an exponential decay in the number of connections with increasing distance.
In other words, they express modular, \emph{small-world}, heavy-tailed, and metabolically constrained organization characteristics.
In small-world networks, most edges form small, densely connected clusters and the others maintain connections between these clusters (see Fig.~\ref{fig:small-world-network}).
This mixture of local clusters and global interaction generates a structure that provides function and integration in the brain that can support a wide range of complex computations, cognition and behavior~\cite{Bullmore_Sporns09}.

By restricting the types of \acp{SNN} that can be implemented in neuromorphic processors to small-world networks, we can dramatically reduce the memory required to specify the routing/connectivity schemes while still supporting a wide range of computations for solving pattern recognition and signal processing tasks.

\subsection{Routing schemes for small-world networks}

Multi-core neuromorphic processors usually use \ac{NOC} designs for managing the communication of neurons between cores, especially asynchronous \acp{NOC}, which are consistent with the event-driven characteristics of neuromorphic processors, avoiding the use of a global clock, thereby reducing power consumption.

Different neuromorphic chips adopt different \ac{NOC} architectures according to the application.
Mesh architectures~\cite{Davies_etal18} represent an easy way to build large-scale systems but at the cost of high routing power and latency.
In flattened butterfly architectures~\cite{Chen_etal19}, neuron cores belonging to the same row and column can communicate directly, with lower routing latency, but with the disadvantages of large area cost and poor multi-casting support.
In~\cite{Park_etal16}, the authors proposed a hierarchical architecture that overcomes some of these disadvantages.
However, it uses off-chip \ac{DRAM} to store the routing lookup table, which significantly increases power consumption.

Current methods for saving power, adopt in- or near-memory computing strategies. However, when on-chip memory is used to store configurable neuron connections 
the required hardware area increases proportionally with the number of neurons and synapses.
For example in the \ac{DYNAP}-SE~\cite{Moradi_etal18}, which uses on-chip hierarchical routing, the routing memory takes around 80\% of the area, even though a combination of point-to-point source-address routing and multicast destination-address routing was used to reduce memory usage.

In the small-world networks that we consider, the number of connections decays with distance, i.e., short-range connections are more common than long-range connections.
Therefore, the strategy we propose to minimize memory requirements is to reduce the address space (i.e., the number of bits) required to map the (many) connections between nearby neurons, and allocate more bits for larger address space domains for the sparse long range connections.

\begin{figure}
  \centering
  \includegraphics[width=0.25\textwidth]{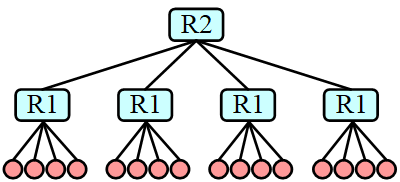}
  \caption{Hierarchical routing scheme optimized for small-world networks.
  Three levels of routers are used: $R0$ to connect neurons inside the same core (red nodes), $R1$ connecting a fixed number of cores, and $R2$ connecting R1 routers.}
	\label{fig:routing}
\end{figure}

Fig.~\ref{fig:routing} shows our hierarchical routing scheme designed to support small-world network connectivity with on-chip memory.
It contains three levels of routers: $R0$ (not explicitly shown) connecting neurons inside the same core (red nodes), $R1$ connecting a fixed number of cores, and $R2$ connecting R1 routers.
Inside a core, there is no specificity: independent of which neuron is sending a spike, all other neurons inside the same core can receive it.
Thus, densely connected clusters are ideal for our core, where a crossbar is used (as $R0$), requiring almost no memory for these connections.
Only a single bit is needed to connect all the neurons in the same core.
We need to use more bits at higher router levels since the address space and specificity increase.
Connectivity between cores depends on the distance between them.
Nearby cores (that use only $R1$ routers) have an increase in specificity, i.e., a subset of neurons in a core can be the target of another core.
At the $R1$ level, each core is divided into two halves, and two rows of one bit each are used to specify connections.
Each neuron in a core can receive spikes from half of the neighboring cores.
At the $R2$ level, connections can be made between neurons below different sets of $R1$ level routers.
Since more neurons can be reached, more bits are necessary to describe the connections.
However, fewer connections exist.
At the $R2$ level, each core is divided into four fourths, so four rows of two bits each are used to specify connections.
These connections have a constrained address space, i.e., since there is an upper bound on the router to which a spike can be sent, it is not necessary to take into account all of the neurons on the whole chip.
In a model going up to the $R2$ level, each neuron needs two one-bit rows and four two-bit rows (i.e., 10 bits in total), allowing a fan-in from up to half of the neurons in the cores that can be reached through the $R1$ level plus a fourth of the neurons in the cores that can be reached through $R2$.

To support this reduction in address space we compute the routing distance by combining the use of computing logic and memory at each router module and update the distance information in the packet, as it transverses the router, thus reducing the address space to the bare minimum needed by the local cluster.

\section{Hardware-software co-design strategies}
\label{sec:hardware-software-codesign}

\begin{figure}
  \centering
  \includegraphics[width=0.35\textwidth]{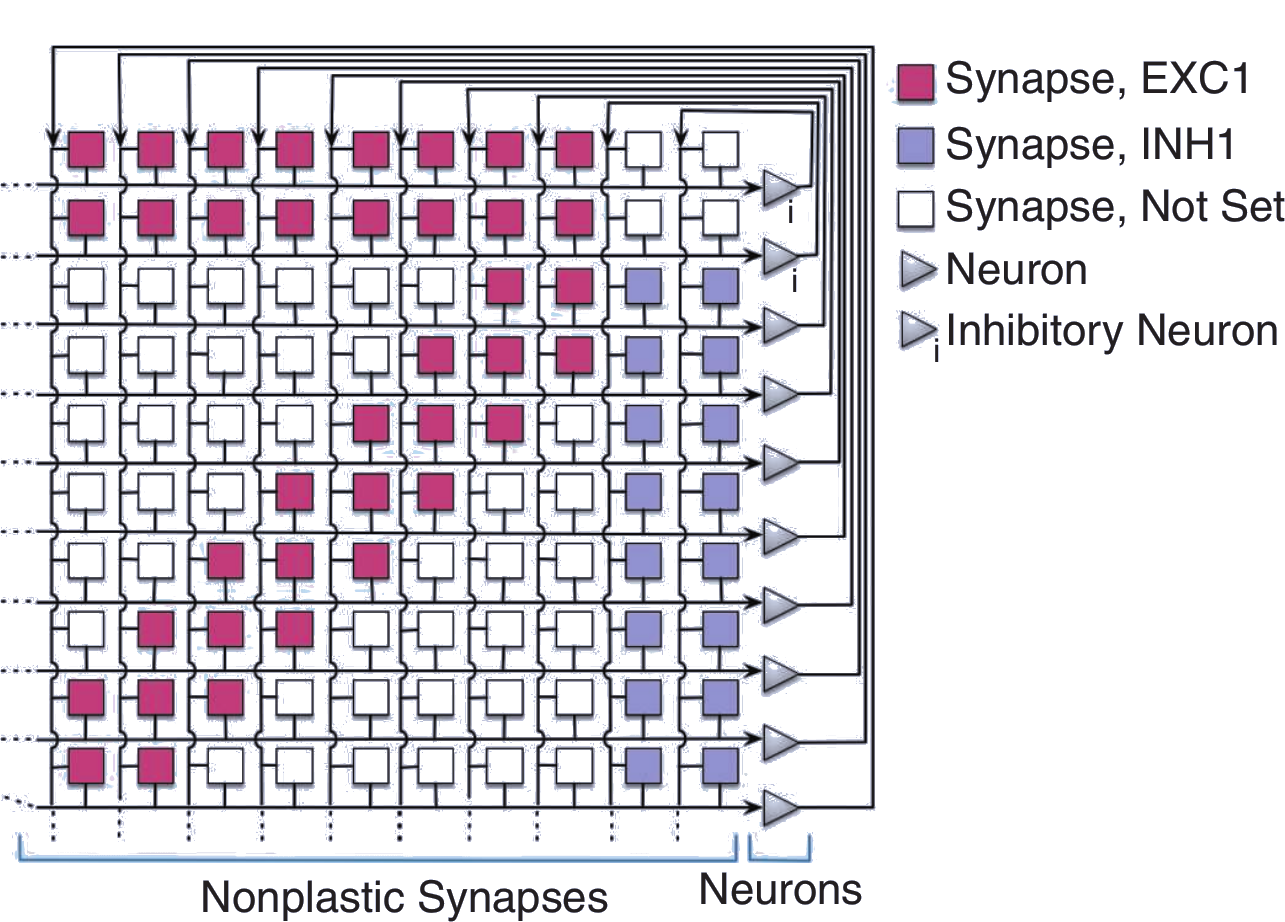}
  \caption{Example of a ``Winner-Takes-All'' (WTA) network implemented on a neuromorphic processor (from~\cite{Indiveri_Sandamirskaya19}). Blue blocks represent inhibitory synapses with negative weights, and red blocks represent excitatory synapses with positive weights.}
  \label{fig:wta-matrix}
\end{figure}

The scheme presented is optimal for \ac{WTA}-like ``canonical''  networks that have a small-world connectivity matrix of the type shown in Fig.~\ref{fig:wta-matrix}.
To map other types of small-world networks in the hardware efficiently (i.e., minimizing memory usage), it is necessary to follow a hardware-software co-design approach, by developing a hardware-aware \emph{neuromorphic compiler}.
A compiler can be seen as a translation machine, i.e., a program that can translate a high-level description of an \ac{SNN} model into the hardware-specific configuration without exposing its fine structure.
By following the hardware-software co-design approach proposed, in addition to placing \ac{SNN} models on existing hardware, the tool developed will also allow us to derive specifications for new chip designs.

\subsection{Hierarchical routing placement example}
Here we present a placement algorithm that maps neurons of a user-defined neural network model onto physical neuron circuits within specific cores of the neuromorphic processor, while adhering to the specificity and distance-based connectivity constraints imposed by the hardware.

\begin{figure}
  \centering
  \includegraphics[width=0.4\textwidth]{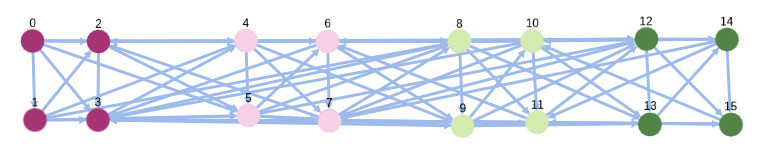}
  \caption{Canonical network example.
           A network with 16 neurons divided into four populations of four neurons each.
           Each population is all-to-all connected.
           The number of connections between populations depends on the distance between them.}
  \label{fig:simple-network-graph-1d-colors}
\end{figure}

\begin{figure}
	\centering
	\subfloat[Canonical network adjacency matrix.
	X- and Y-axis enumarate the neurons in the network.
	Zero means no connection between neurons, and one means a connection from the $i$-th neuron in the row to the $j$-th neuron in the column.
	There is no self connection in our network, thus the main diagonal is all zeros.
	\label{fig:canonical-adj-matrix}]{\includegraphics[width=0.22\textwidth]{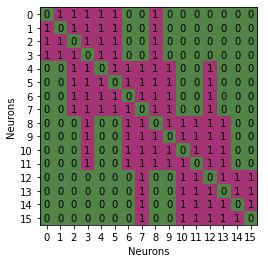}}\hfill
	\subfloat[Number of connections between cluster of neurons (all-to-all connected).
	These clusters define the cores in our proposed architecture.
	X- and Y-axis enumerate the clusters.
	Note how the number of connections between cores decay with distance.\label{fig:canonical-number-connections}]{\includegraphics[width=0.21\textwidth]{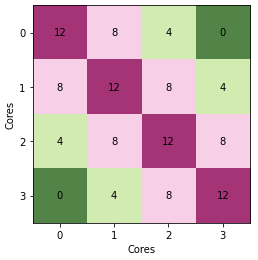}}\hfill
	\caption{Connectivity matrices.}
	\label{fig:canonical-connectivity}
\end{figure}

To validate the placement algorithm we test it with a canonical network, as depicted in Fig.~\ref{fig:simple-network-graph-1d-colors},  and a hypothetical neuromorphic processor comprising four cores, with four neurons per core.
The canonical network considered has four populations of four neurons, where each population is all-to-all connected (Fig,~\ref{fig:canonical-adj-matrix}, and the number of connections between populations decays with distance. Figure~\ref{fig:canonical-number-connections} shows the number of connections between neurons and populations.

\subsection{Placement procedure}

To place the network, we follow three main steps:
(i) place neurons in cores given the connections among them;
(ii) create a map of distances among the cores created;
(iii) place the connections between the neurons.

\begin{figure}
	\centering
	\includegraphics[width=0.3\textwidth]{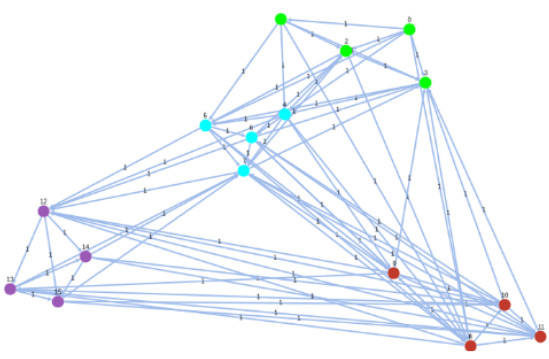}
	\caption{Canonical network placement.
          By finding the cliques, we place the neurons in the same core.
          Different colors denote different cores.}
	\label{fig:placing-nodes}
\end{figure}

To place neurons in cores, we find \emph{cliques} (i.e., the sets of neurons that have the largest number of connections among each other) in the graph of the network.
Each clique defines a core.
This gives us the number of cores needed to place the network.
Figure~\ref{fig:placing-nodes} shows the result of grouping the neurons in the canonical network with this method.

With cores defined, we can calculate the distance between cores $i$ and $j$, with $i \neq j$ as:
\[
\mathit{dist}[i][j] = \mathit{floor}((n/e(i,j)))+1
\]
where $n$ is the number of neurons per core, and $e(i,j)$ is the number of connections core $i$ is receiving from core $j$.
Note that our distance is a quasi-metric that can be non-symmetric, i.e., the distance from core $i$ to $j$ might not be the same as from $j$ to $i$.
The maximum distance in the network indicates the maximum router depth we need to use to map the network.
We define the distance between the core and itself ($i=j$) as zero, and between two cores that do not share connections as --1.
Having determined the number of cores and distances between cores, we can place connections, starting with the closest pairs.
Each neuron has a limited set of synapses;
we try to allocate the nearby connections first because they are more numerous than the further-away ones.
If no synapse is available, the connection is flagged as unplaceable.
We consider the option of having a few fully programmable synapses to accommodate flagged connections.

\section{Preliminary Results}

With this heuristic, using our tailor-made canonical network, we verify that the algorithm places the neurons correctly.
To demonstrate how the algorithm performs, we test it using networks that deviate from the canonical example shown in Fig.~\ref{fig:simple-network-graph-1d-colors} created by removing an increasing number of nodes.

\begin{figure}
	\centering
	\includegraphics[width=0.4\textwidth]{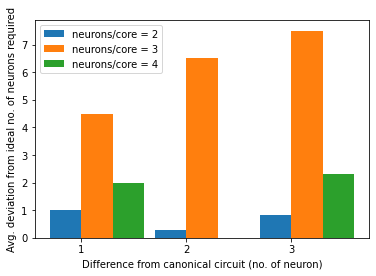}
	\caption{Placement algorithm result showing on average the extra number of hardware neurons required to map three new models that deviate from the canonical network.
          The three new models were generated from the canonical network by removing one, two, or three neurons (x-axis).
          For each model different combinations of networks with the same number of neurons are generated.
          The total number of hardware neurons required to map network models depend on the design choice of core size (i.e., the number of neurons allocated per core).
          Three bars are plotted for cores of size 2 (blue), 3 (orange), and 4 (green).
        }
	\label{fig:preliminary-results}
\end{figure}

Figure~\ref{fig:preliminary-results} compares the results of the placement algorithm deviation for three different network variations with one, two and three fewer nodes, and for cores of different size.
Our canonical network was designed based on four cores of four neurons, but we can see that also cores with two neurons can map the network efficiently, with a small overhead (and even smaller than using cores of size four).
Also, we can see how the average number of neurons required in the hardware increases when using three neurons per core, making it a not suitable option.

\section{Conclusion}
The development of domain-specific neuromorphic hardware can help to advance \ac{AI} for edge-computing taks, and our understanding of to optimize resource allocation in  brain-line small-world networks.
In this work, we presented a hardware-software co-design approach, where a heuristic for placing an \ac{SNN} into neuromorphic hardware also gives direction to guide new chip designs (e.g., specifying how many neurons/core, how many cores, and how many router hierarchy levels to use).
The automatic mapping of networks done following biological constraints, and the specifications it provides for  designing  new multi-core \ac{SNN} chips will allow us to best exploit future generations of neuromorphic processors.
Our placement algorithm gives us some options for optimal hardware parameters for a specific set of network structures.
We obtain different viable options for a chip design in terms of the number of neurons per core, cores, and synapses, which can be used to calculate the real area/memory necessary to design a new chip.
This way, our co-design approach allows us to reduce the search space of new hardware parameters.

%

\bibliographystyle{ieeetr}
\bibliography{biblio/biblioncs}

\end{document}